\documentclass[11pt,a4paper]{article}
\usepackage[dvipsnames]{xcolor}
\usepackage[hyperref]{naaclhlt2019}
\usepackage{times}
\usepackage{latexsym}
\usepackage[normalem]{ulem}
\usepackage{enumitem}
\usepackage{url}

\usepackage{graphicx}
\usepackage{booktabs}
\usepackage{tabularx}
\usepackage{cleveref}
\usepackage{todonotes}
\usepackage[toc,page]{appendix}
\usepackage{xspace}
\usepackage{subfig}
\usepackage{framed}

\newcommand{\appref}[1]{Appendix~\ref{sec:#1}}
\newcommand{\labeledsection}[2]{\section{#1}\label{sec:#2}}

\newcommand{\term}[0]{challenge\xspace}
\newcommand{\Term}[0]{Challenge\xspace}

\aclfinalcopy % Uncomment this line for the final submission
\def\aclpaperid{1951} %  Enter the acl Paper ID here

\title{Inoculation by Fine-Tuning: A Method for Analyzing \Term Datasets}

\author{ 
    Nelson F. Liu$^{\spadesuit\heartsuit\clubsuit}$ \quad
	Roy Schwartz$^{\spadesuit\clubsuit}$ \quad 
	{\bf Noah A.~Smith}$^{\spadesuit\clubsuit}$ \\
    $^\spadesuit$Paul G. Allen School of Computer Science \& Engineering,\\
    University of Washington, Seattle, WA, USA \\
	$^\heartsuit$Department of Linguistics, University of Washington, Seattle, WA, USA \\
	$^\clubsuit$Allen Institute for Artificial Intelligence, Seattle, WA, USA \\
	{\tt \{nfliu,roysch,nasmith\}@cs.washington.edu}}

\date{}

\begin{document}
\maketitle
\begin{abstract}
Several datasets have recently been  constructed to expose brittleness in models trained on existing benchmarks.
While model performance on these \textit{\term datasets} is significantly lower compared to the original benchmark, it is unclear what particular weaknesses they reveal.
For example, a \term dataset may be difficult because it targets phenomena that current models cannot capture, or because it simply exploits blind spots in a model's specific training set.
We introduce {\it inoculation by fine-tuning}, a new analysis method for studying \term datasets by exposing models (the metaphorical patient) to a small amount of data from the \term dataset (a metaphorical pathogen) and assessing how well they can adapt.
We apply our method to analyze the NLI ``stress tests'' \cite{Naik2018StressTE} and the Adversarial SQuAD dataset \cite{Jia2017AdversarialEF}. 
We show that after slight exposure, some of these datasets are no longer challenging, while others remain difficult. Our results indicate that failures on \term datasets may lead to very different conclusions about models, training datasets, and the \term datasets themselves.

\end{abstract}

\section{Introduction}
NLP research progresses  through the construction of dataset-benchmarks and the development of systems whose performance on them can be fairly compared.
A recent pattern involves \emph{challenges} to benchmarks:\footnote{Often referred to as ``adversarial datasets'' or ``attacks''.}  manipulations to input data that result in severe degradation of system performance, but not human performance.
These challenges have been used as evidence that current systems are brittle \cite[\emph{inter alia}]{Belinkov:2018,Mudrakarta:2018,Zhao:2018,Glockner2018BreakingNS,Ebrahimi:2018,Ribeiro:2018}.
For instance, \citet{Naik2018StressTE} generated natural language inference \term data by applying simple textual transformations to existing examples from MultiNLI \citep{Williams2018ABC} and SNLI \citep{Bowman2015ALA}. Similarly, \citet{Jia2017AdversarialEF}  built an adversarial evaluation dataset for reading comprehension based on SQuAD \citep{Rajpurkar2016SQuAD10}.

\begin{figure}
	\centering
	\includegraphics[width=0.85\columnwidth]{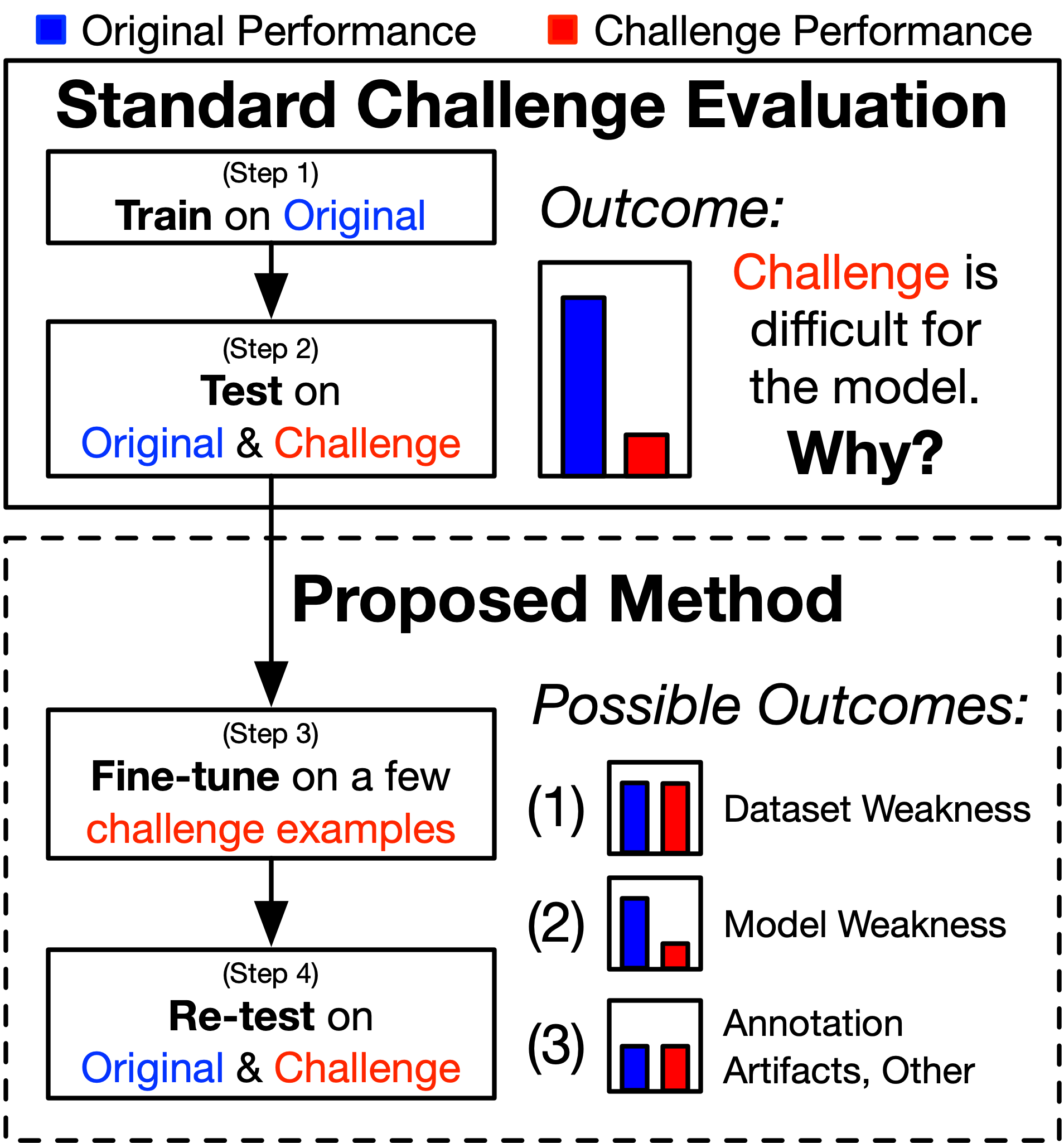}
	\caption{An illustration of the standard \term evaluation procedure \citep[e.g.,][]{Jia2017AdversarialEF} and our proposed analysis method. {\color{blue} ``Original''} refers to the a standard dataset (e.g., SQuAD) and {\color{red} ``\Term''} refers to the \term dataset (e.g., Adversarial SQuAD).
	Outcomes are discussed in Section~\ref{sec:method}.
  \label{fig:method_summary}}
\end{figure}

What should we conclude when a system fails on a \term dataset?
In some cases, a \term might exploit blind spots in the design of the \emph{original} dataset (\emph{dataset weakness}).
In others, the \term might expose an inherent inability of a particular model family to handle certain natural language phenomena (\emph{model weakness}). These are, of course, not mutually exclusive.

We introduce \textbf{inoculation by fine-tuning}, a new method for analyzing the effects of \term datasets (\Cref{fig:method_summary}).\footnote{\emph{Inoculation} evokes the idea that treatable diseases have different implications (for society and for the patient) than untreatable ones.  We differentiate the abstract process of inoculation from our way of executing it (fine-tuning) since it is easy to imagine alternative ways to inoculate a model.}
Given a model trained on the original dataset, we expose it to a small number of examples from the \term dataset, allowing learning to continue.
To the extent that the weakness lies with the original dataset, then the inoculated model will perform well on both the original and \term held-out data (Outcome 1 in \Cref{fig:method_summary}).
If the weakness lies with the model, then inoculation will prove ineffective and the model's performance will remain unchanged (Outcome 2).

Inoculation can also decrease a model's performance on the original dataset (Outcome 3).  
This case is not as clear as the first two, and could result from systematic differences between the original and \term datasets, due to, e.g., predictive artifacts in either dataset \cite{Gururangan2018AnnotationAI}.

We apply our method to analyze six \term datasets: the \textit{word overlap}, \textit{negation}, \textit{spelling errors}, \textit{length mismatch} and  \textit{numerical reasoning} NLI challenge datasets proposed by \citet{Naik2018StressTE}, as well as the Adversarial SQuAD reading comprehension challenge dataset \cite{Jia2017AdversarialEF}.
We analyze NLI datasets with the ESIM \citep{Chen2017EnhancedLF} and the decomposable attention \cite{Parikh2016ADA} models, and reading comprehension with the BiDAF \citep{Seo2016BidirectionalAF} and the QANet \citep{Yu2018QANetCL} models.

By fine-tuning on, in some cases,  as few as 100 examples, both NLI models are able to recover almost the entire performance gap on both the \emph{word overlap} and \emph{negation} challenge datasets (Outcome 1).
In contrast, both models struggle to adapt to the \emph{spelling error} and \emph{length mismatch} challenge datasets (Outcome 2). 
On the \emph{numerical reasoning} challenge dataset, both models close all of the gap using a small number of samples, but at the expense of performance on the original dataset (Outcome 3).
For Adversarial SQuAD, BiDAF closes 60\% of the gap with minimal fine-tuning, but suffers a 7\% decrease in original test set performance (Outcome 3). QANet shows similar trends.

Our proposed analysis is broadly applicable, easy to perform, and task-agnostic.
By gaining a better understanding of how challenge datasets stress models, we can better tease apart limitations of datasets and limitations of models.

\begin{table*}
\centering
\setlength\tabcolsep{4pt}
\footnotesize
\begin{tabularx}{\linewidth}{lXX}
\toprule
Category & Premise & Hypothesis \\
\midrule
Word Overlap & Possibly no other country has had such a turbulent history. & The country's history has been turbulent \textbf{and true is true}.\\
\midrule
Negation & Possibly no other country has had such a turbulent history. & The country's history has been turbulent \textbf{and false is not true}.\\
\midrule
Spelling Errors & I have done what you asked. & I have disobeyed your \textbf{ordets}. \\
\midrule
Length Mismatch & Possibly no other country has had such a turbulent history \textbf{and true is true and true is true and true is true and true is true and true is true}. & The country's history has been turbulent.\\
\midrule
Numerical Reasoning & \textbf{Tim has 350 pounds of cement in 100, 50, and 25
pound bags.} & \textbf{Tim has less than 750 pounds of cement in 100, 50, and 25 pound bags.} \\
\bottomrule
\end{tabularx}
\caption{Examples from each of the NLI challenge datasets analyzed, a subset of a broader suite of NLI stress tests proposed by \citet{Naik2018StressTE}. Boldface denotes perturbations to original MultiNLI examples. Figure contents reproduced from \citet{Naik2018StressTE}.}
\label{tab:nli_stress_test_examples}
\end{table*}

\section{Inoculation by Fine-Tuning} 
\label{sec:method}

Our method assumes access to an original dataset divided into training and test portions, as well as a challenge dataset, divided into a (small) training set\footnote{The exact amount of challenge data used for fine-tuning might affect our conclusions, so we consider different sizes of the ``vaccine'' in our experiments.} and a test set.  After training on the original (training) data, we measure system performance on both test sets. 
We assume the usual observation---a generalization gap with considerably lower performance on the challenge test set.

We then proceed to fine-tune the model on the challenge training data, i.e., continuing to train the pre-trained model on the new data until development performance on the original development set has not improved for five epochs.\footnote{The use of the original development set is meant to both prevent us from using more challenge data and verify that the learner does not completely forget the original dataset.}
Finally, we measure performance of the inoculated model on both the original and challenge test sets.
Three clear outcomes of interest are:\footnote{The outcome may also lie between these extremes, necessitating deeper analysis.}

\paragraph{Outcome 1} The gap closes, i.e., the inoculated system retains its (high) performance on the original test set and performs as well (or nearly so) on the challenge test set.  
This case suggests that the challenge dataset did not reveal a weakness in the  model family.  Instead, the challenge has likely revealed a lack of diversity in the original dataset.
\paragraph{Outcome 2} Performance on both test sets is unchanged. 
This indicates that the challenge dataset has revealed a fundamental weakness of the model; it is unable to adapt to the challenge data distribution, even with some exposure.
\paragraph{Outcome 3} Inoculation damages performance on the original test set (regardless of improvement on the challenge test set).
The main difference between Outcome 3 and Outcomes 1 and 2 is that here, by fine-tuning, the model is shifting towards a challenge distribution that somehow contradicts the original distribution. 
This could result from, e.g., a different label distribution between both datasets, or annotation artifacts that exist in one dataset but not in the other (see  Sections~\ref{sec:results}, \ref{sec:discussion}).

\begin{figure}[!htbp]
\small
\begin{framed}
\footnotesize
  \textbf{Article:} Super Bowl 50

  \textbf{Paragraph:}
  \emph{Peyton Manning became the first quarterback ever to lead two different teams to multiple Super Bowls. He is also the oldest quarterback ever to play in a Super Bowl at age 39. The past record was held by John Elway, who led the Broncos to victory in Super Bowl XXXIII at age 38 and is currently Denver's Executive Vice President of Football Operations and General Manager.
  \textcolor{blue}{Quarterback Jeff Dean had jersey number 37 in Champ Bowl XXXIV.}}

  \textbf{Question:} \emph{What is the name of the quarterback who was 38 in Super Bowl XXXIII?}

\end{framed}
\caption{
An example from the Adversarial SQuAD dataset, with the distractor sentence \textcolor{blue}{in blue}. Figure reproduced from \citet{Jia2017AdversarialEF}.
}
\label{fig:adversarial_squad_example}
\end{figure}

\section{Not all Challenge Datasets are Alike}

To demonstrate the utility of our method, we apply it to analyze the NLI stress tests  \cite{Naik2018StressTE} and the Adversarial SQuAD dataset \cite{Jia2017AdversarialEF}. We fine-tune models on a varying number of examples from the challenge dataset training split in order to study whether our method is sensitive to the level of exposure.\footnote{See \appref{experimental_setup_details} for experimental process details.}
Our results demonstrate that different challenge datasets lead to different outcomes.
We release code for reproducing our results.\footnote{\url{http://nelsonliu.me/papers/inoculation-by-finetuning}}

\begin{figure*}[h]
\begin{tabular}{ccc}
\textbf{\Large Outcome 1} &  \textbf{\Large Outcome 2} & \textbf{\Large Outcome 3} \\\\
\textbf{(a) Word Overlap} & \textbf{(c) Spelling Errors} & \textbf{(e) Numerical Reasoning} \\ 
\includegraphics[width=0.3\linewidth]{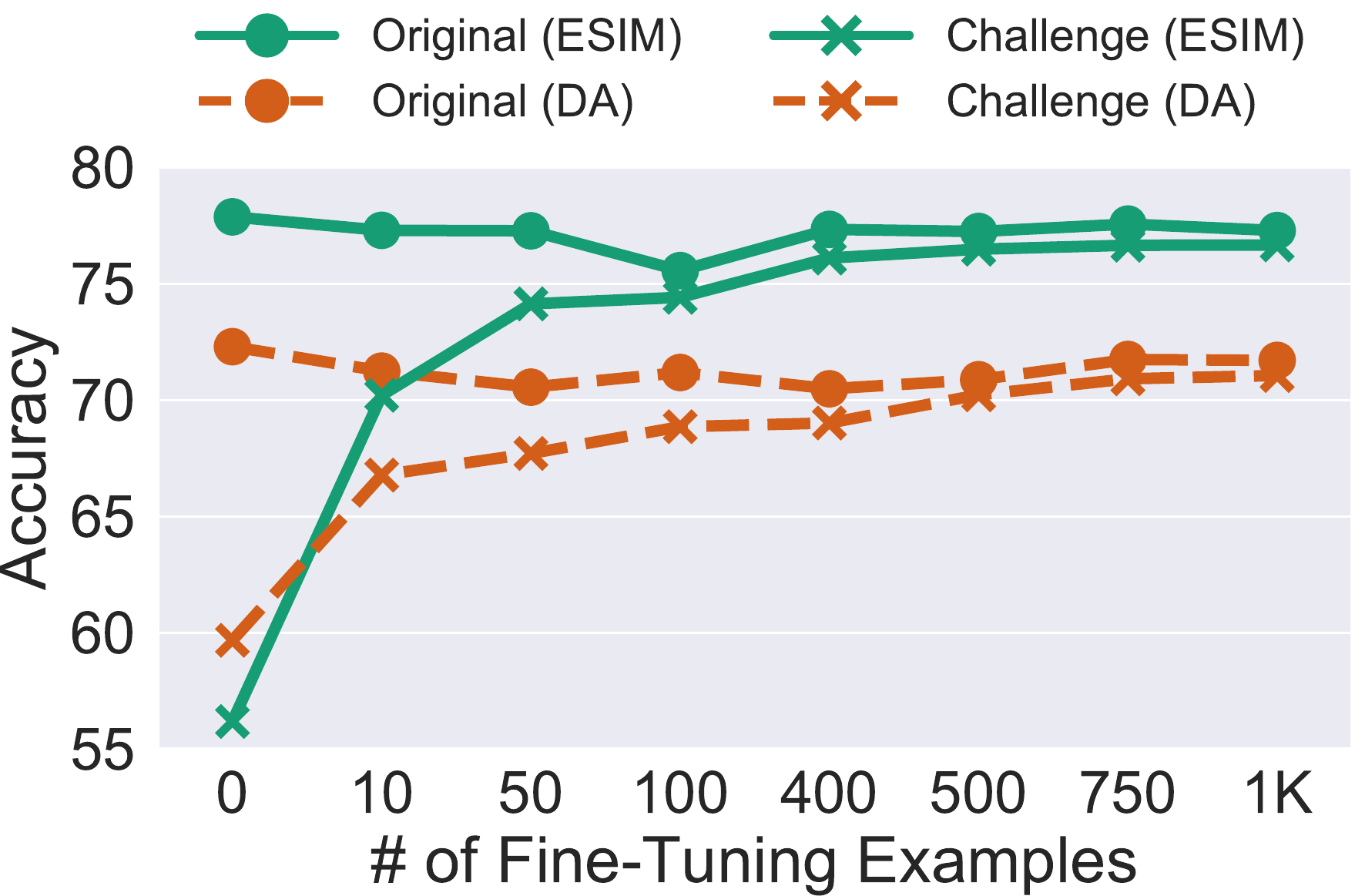} &
 {\subfloat{\includegraphics[width=0.3\linewidth]{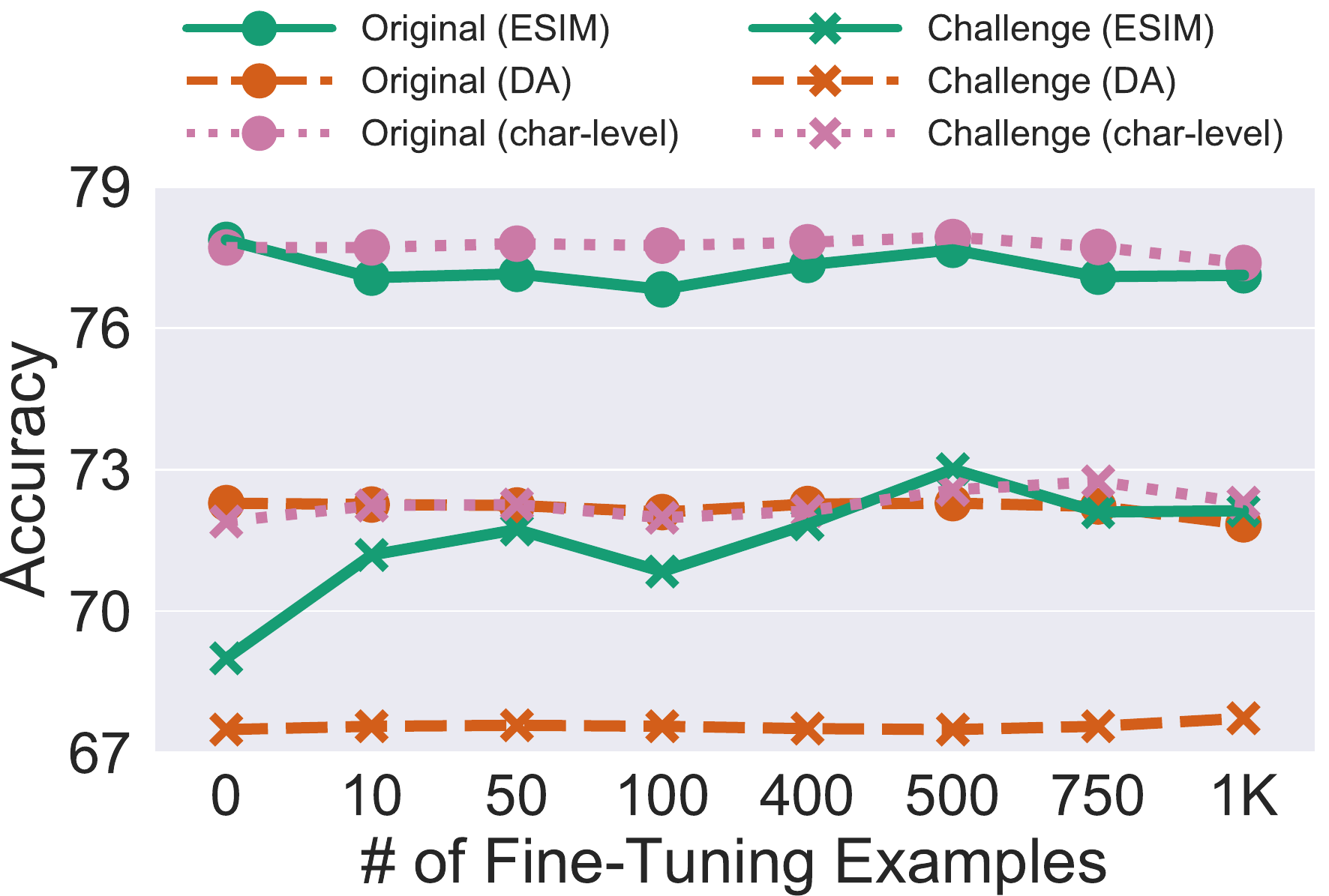}}} &
 {\subfloat{\includegraphics[width=0.3\linewidth]{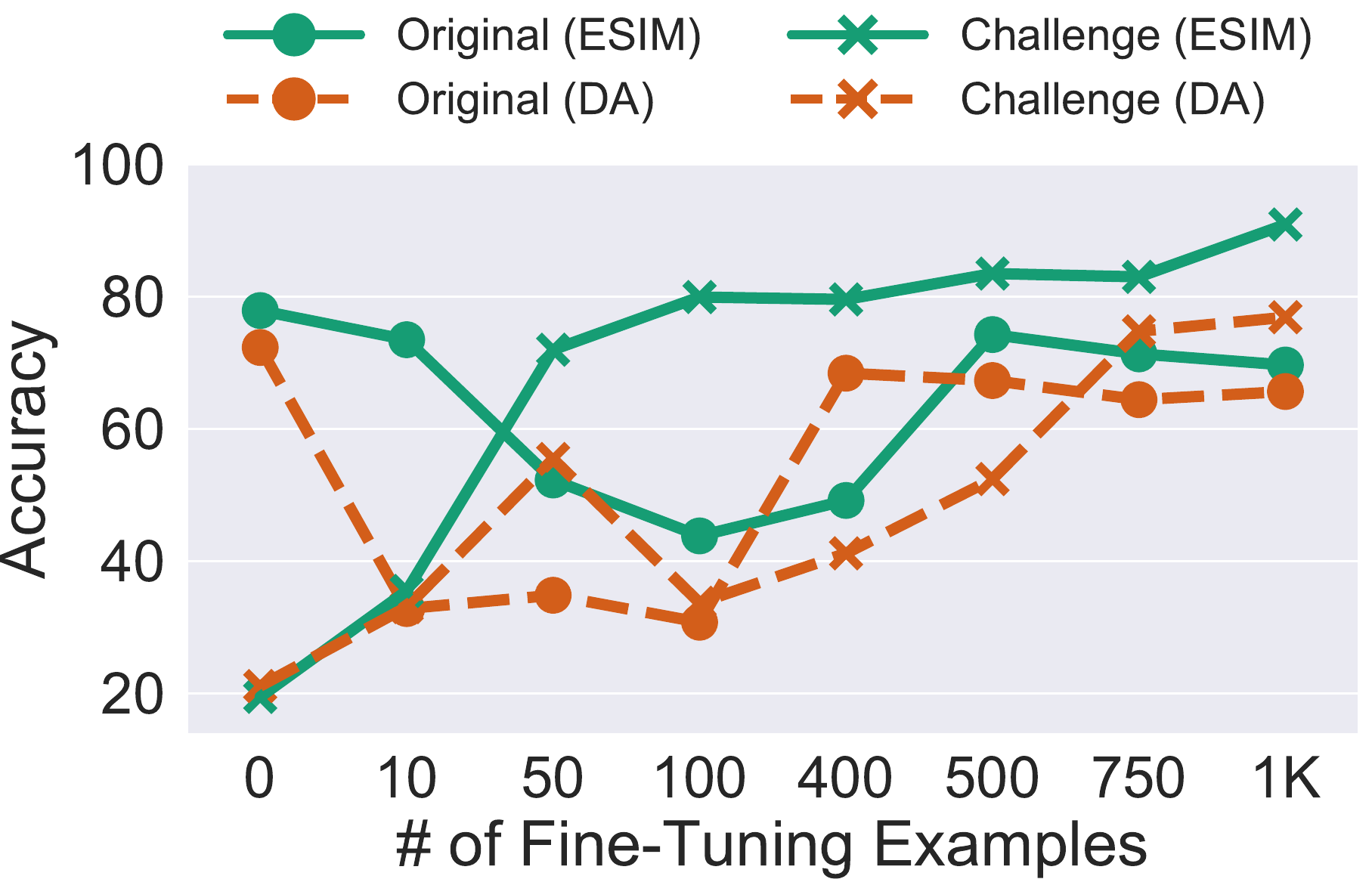}}} \\ \\
{\textbf{(b) Negation}} & {\textbf{(d) Length Mismatch}} & \textbf{(f) Adversarial SQuAD}\\
 \subfloat{\includegraphics[width=0.3\linewidth]{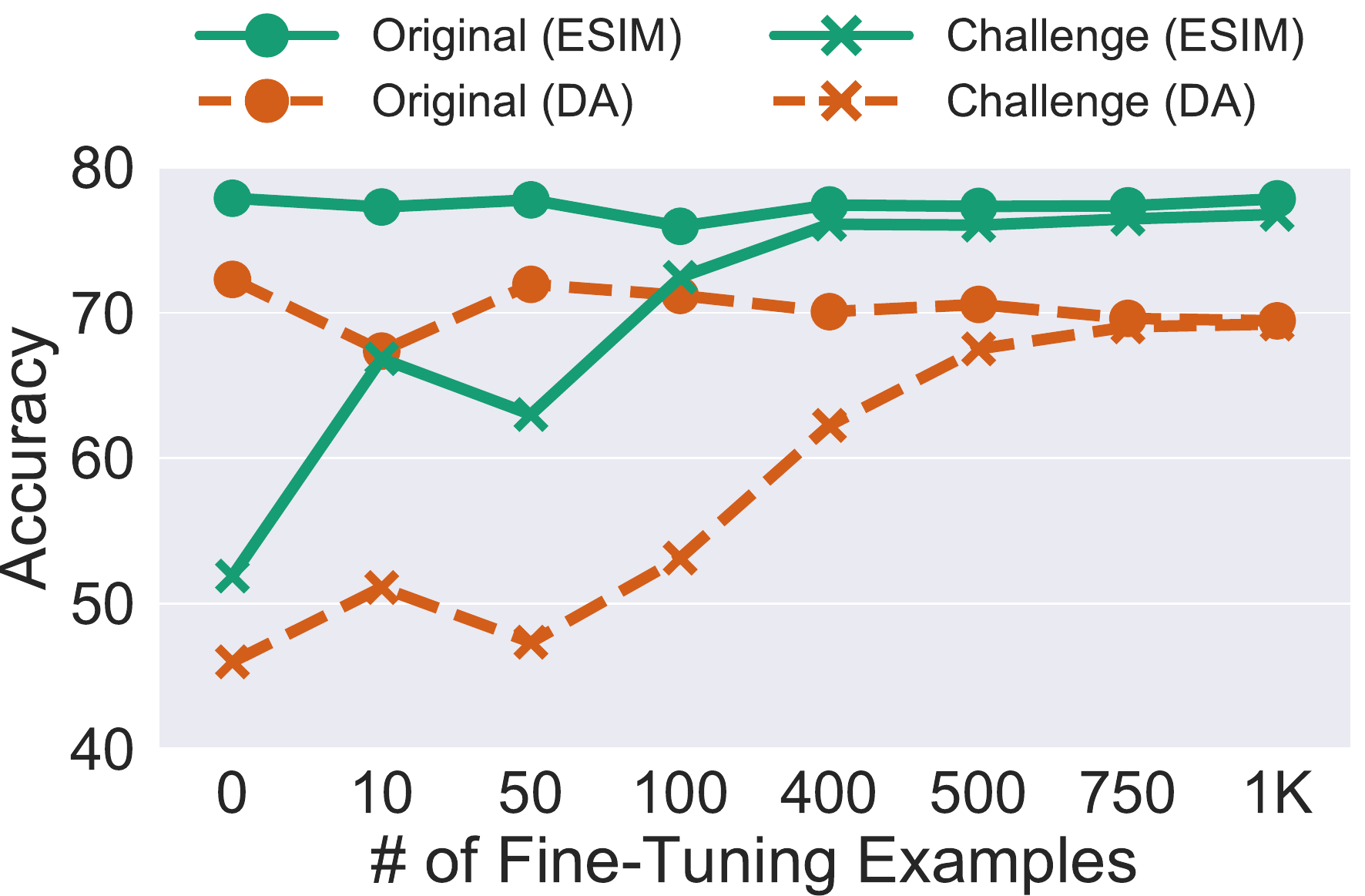}} & 
{\subfloat{\includegraphics[width=0.3\linewidth]{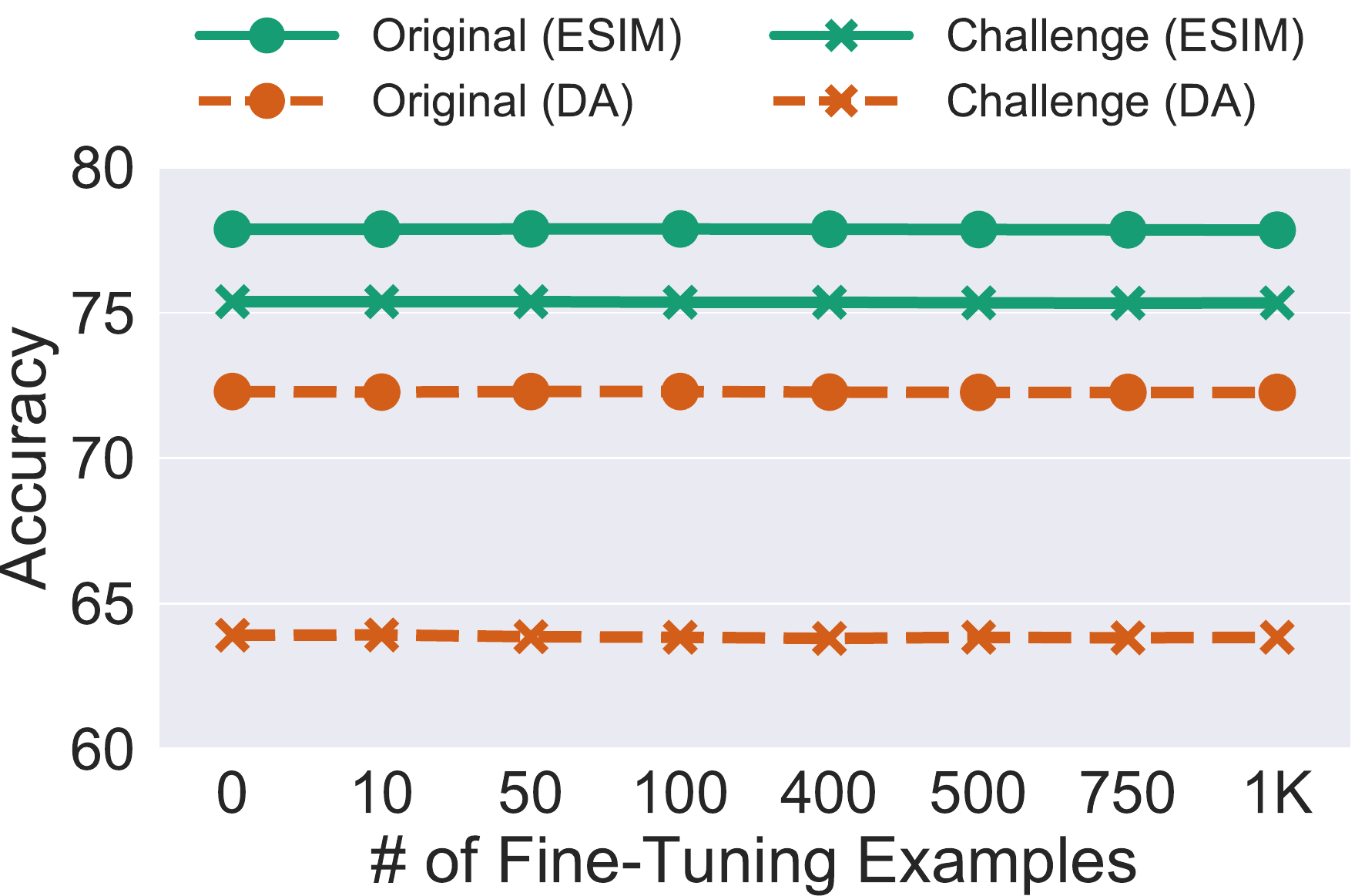}}} &
\includegraphics[width=0.3\linewidth]{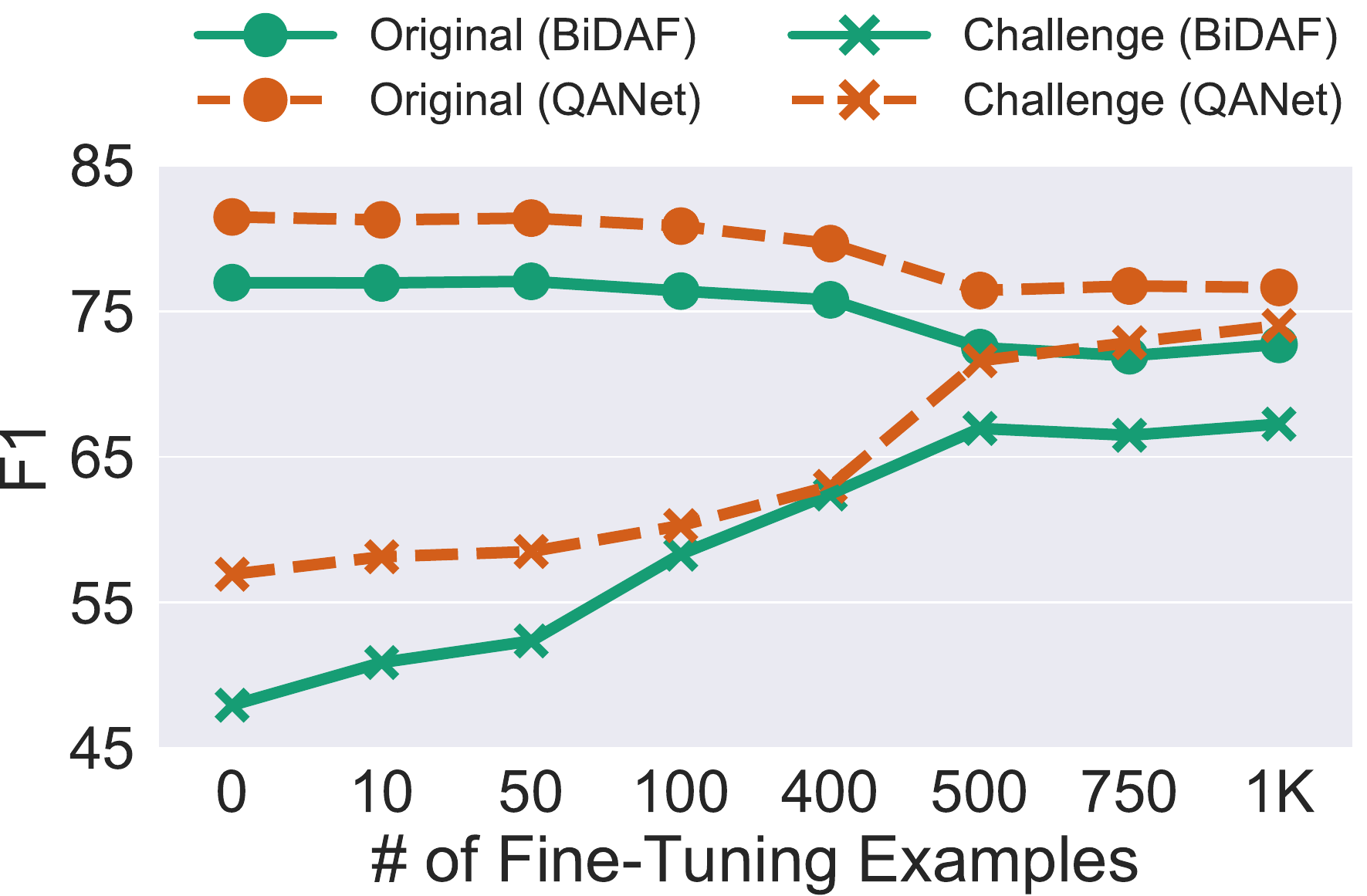}
\end{tabular}
\caption{Inoculation by fine-tuning results. 
(a--e): NLI accuracy for the ESIM and decomposable attention (DA) models.
(f): Reading comprehension $F_1$ scores for the BiDAF and QANet models.\\
Fine-tuning on a small number of word overlap (a) and negation (b) examples erases the performance gap (Outcome 1). Fine-tuning does not yield significant improvement on spelling errors (c) and length mismatch (d), but does not degrade original performance either (Outcome 2). Fine-tuning on numerical reasoning (e) closes the gap entirely, but also reduces performance on the original dataset (Outcome 3).
On Adversarial SQuAD (f), around 60\% of the performance gap is closed after fine-tuning, though performance on the original dataset decreases (Outcome 3).
On each challenge dataset, we observe similar trends between different models. 
}
  \label{fig:results}
\end{figure*}

\subsection{Datasets}

We briefly describe the analyzed datasets, but refer readers to the original publications for details.

\paragraph{NLI Stress Tests} \citet{Naik2018StressTE} proposed six automatically-constructed ``stress tests'', each focusing on a different weakness of NLI systems. We analyze five of these stress tests (\Cref{tab:nli_stress_test_examples}).\footnote{The remaining  challenge dataset---\textbf{antonym}---is briefly discussed in Section~\ref{sec:discussion}.}

The \textbf{word overlap} challenge dataset is designed to exploit models' sensitivity to high lexical overlap in the premise and hypothesis by appending the tautology \textit{``and true is true''} to  the hypothesis. The \textbf{negation}  challenge dataset is based on the observation that negation words (e.g., \textit{``no''}, \textit{``not''}) cause the model to classify neutral or entailed statements as contradiction. In this dataset, the tautology \textit{``and false is not true''} is appended to the hypothesis sentence.
The \textbf{spelling errors}  challenge dataset is designed to evaluate model robustness to noisy data in the form of misspellings. 
The \textbf{length mismatch} challenge dataset is designed to exploit models' inability to handle examples with much longer premises than hypotheses. In this dataset, the tautology \textit{``and true is true''} is appended five times to the end of the premise.
Lastly,  the \textbf{numerical reasoning}  challenge dataset is designed to test models' ability to perform algebraic calculations, by introducing premise-hypothesis pairs containing numerical expressions. 

We analyze these challenge datasets using two models, both trained on the MultiNLI dataset:\footnote{MultiNLI has domain-matched and mismatched development data, so we train separate ``matched'' and ``mismatched'' models that each use the corresponding development set for learning rate scheduling and early stopping.
We observe similar results in both cases, so we focus on the models trained on ``matched'' data. See \appref{mismatched_results} for mismatched results.} the ESIM model \citep{Chen2017EnhancedLF} and the decomposable attention model (DA; \citealp{Parikh2016ADA}).

To better address the spelling errors challenge dataset, we also train a character-sensitive version of the ESIM model.
We concatenate the word representations with the 50-dimensional hidden states produced by running each token through a character bidirectional GRU \cite{Cho:2014}.

\paragraph{Adversarial SQuAD} \citet{Jia2017AdversarialEF} created a challenge dataset for reading comprehension  by appending automatically-generated distractor sentences to SQuAD passages. 
The appended distractor sentences are crafted to look similar to the question while not contradicting the correct answer or misleading humans (\Cref{fig:adversarial_squad_example}). 
The authors released model-independent Adversarial SQuAD examples, which we analyze. 
For our analysis, we use the BiDAF model \citep{Seo2016BidirectionalAF} and the QANet model \citep{Yu2018QANetCL}.

\subsection{Results}
\label{sec:results}

We refer to difference between a model's pre-inoculation performance on the original test set and the challenge test set as the \textit{performance gap}.

\paragraph{NLI Stress Tests} \Cref{fig:results} presents NLI accuracy for the ESIM and DA models on the word overlap, negation, spelling errors, length mismatch and numerical reasoning challenge datasets after fine-tuning on a varying number of challenge examples.

For the word overlap and negation challenge datasets, both ESIM and DA quickly close the performance gap when fine-tuning (Outcome 1). For instance, on both of the aforementioned challenge datasets, ESIM requires only 100 examples to close over 90\% of the performance gap while maintaining high performance on the original dataset.
Since these performance gaps are closed after seeing a few challenge dataset examples ($<$ 0.03\% of the original MultiNLI training dataset), these challenges are likely difficult because they exploit easily-recoverable gaps in the models' training dataset rather than highlighting their inability to capture semantic phenomena.

In contrast, on spelling errors and length mismatch, fine-tuning does not allow either model to close a substantial portion of the performance gap, while performance on the original dataset is unaffected (Outcome 2).\footnote{The length mismatch dataset is not particularly challenging for the ESIM model: its untuned performance on the challenge set is only 2.5\% lower than its original performance. Nonetheless, this gap remains fixed even after fine-tuning}
Interestingly, the character-aware ESIM model trained on spelling errors shows a similar trend, suggesting that the this challenge set is highlighting a weakness of ESIM that goes beyond the word representation.

On numerical reasoning, the entire gap is closed by fine-tuning ESIM on 100 examples, or DA on 750 examples. However, both models' original dataset performance substantially decreases (Outcome 3; see discussion in Section~\ref{sec:discussion}).

\paragraph{Adversarial SQuAD}

\Cref{fig:results}(f) shows BiDAF and QANet results after fine-tuning on a varying number of challenge samples.

Fine-tuning BiDAF on only 400 challenge examples closes more than 60\% of the performance gap, but also results in substantial performance loss on the original SQuAD development set; fine-tuning QANet yields the same trend (Outcome 3).
In this case, the model likely takes advantage of the fact that the adversarial distractor sentence is always concatenated to the end of the paragraph.\footnote{Indeed, \citet{Jia2017AdversarialEF} show that models trained on Adversarial SQuAD are able to overcome the adversary by simply learning to ignore the last sentence of the passage.}

\subsection{Discussion}
\label{sec:discussion}

\paragraph{Explaining the Numerical Reasoning Results}
The relative ease with which the ESIM model overcomes the numerical reasoning challenge seems to contradict the findings of \citet{Naik2018StressTE}, who observed that ``\emph{the model is unable to perform
reasoning involving numbers or quantifiers \dots}''.
Indeed, it seems unlikely that a model will learn to perform algebraic numerical reasoning based on as few as 50 NLI examples. 

However, a closer look at this dataset provides a potential explanation for this finding. The dataset was constructed such that a simple 3-rule baseline is able to surpass 80\% on the task (see \appref{numeric_explanation}). For instance, 35\% of the dataset examples contain the phrase ``more than'' or ``less than'' in their hypothesis, and 95\% of these have the label ``neutral''. As a result, learning a handful of these rules is sufficient for achieving high performance on this challenge dataset. 

This observation highlights a key property of Outcome 3: 
challenge datasets that are easily recoverable by our method, at the expense of performance on the original dataset, are likely not testing the full breadth of a linguistic phenomenon but rather a specific aspect of it.

\paragraph{Limitations of Our Method}
Our inoculation method assumes a somewhat balanced label distribution in the challenge dataset training portion. 
If a challenge dataset is highly skewed to a specific label, fine-tuning will result in simply learning to predict the majority label; such a model would achieve high performance on the challenge dataset and low performance on the original dataset (Outcome 3). 
For such datasets, the result of our method is not very informative.\footnote{For instance, the \emph{antonym}  challenge dataset of \citet{Naik2018StressTE}, in which all examples are labeled ``contradiction''.}
Nonetheless, as in the numerical reasoning case discussed above, 
this lack of diversity signals a somewhat limited phenomenon captured by the challenge dataset.

\section{Conclusion}

We presented a method for studying \textit{why} challenge datasets are difficult for models.
Our method fine-tunes models on a small number of challenge dataset examples.
This analysis yields insights into models, their training datasets, and the challenge datasets themselves.
We applied our method to analyze the challenge datasets of \citet{Naik2018StressTE} and \citet{Jia2017AdversarialEF}.
Our results indicate that some of these challenge datasets break models by exploiting blind spots in their training data, while others may challenge more fundamental weaknesses of model families. 

\section*{Acknowledgments}

We thank Aakanksha Naik and Abhilasha Ravichander for generating NLI stress test examples from the MultiNLI training split, and Robin Jia for answering questions about the Adversarial SQuAD dataset. We also thank the members of the Noah's ARK group at the University of Washington, the researchers at the Allen Institute for Artificial Intelligence, and the anonymous reviewers for their valuable feedback. NL is supported by a Washington Research Foundation Fellowship and a Barry M. Goldwater Scholarship. This work was supported in part by a hardware gift from NVIDIA Corporation.

\bibliography{naaclhlt2019}
\bibliographystyle{acl_natbib}

\newpage
\clearpage

\begin{appendices}

\labeledsection{Experimental Setup Details}{experimental_setup_details}

\paragraph{Generating challenge training sets}
When varying the size of the challenge dataset train split used for fine-tuning, we subsample inclusively. For example, the dataset used for fine-tuning on 5 examples is a subset of the dataset used for fine-tuning on 100 examples, which is a subset of the dataset used for fine-tuning on 1000 examples.

The \emph{word overlap}, \emph{negation}, \emph{spelling errors} and \emph{length mismatch} NLI challenge datasets, as well as Adversarial SQuAD, include splits for training and evaluation. To generate the datasets used for fine-tuning, we subsample 1000 random examples from each of the challenge dataset train splits.\footnote{For Adversarial SQuAD, we subsample from distinct passages.} The evaluation splits are used as-is.

The \emph{numerical reasoning} NLI challenge dataset is unsplit. As a result, we generate the datasets used for fine-tuning by subsampling 1000 random examples from the entirety of the challenge dataset, and use the remaining examples for evaluation.

\paragraph{Experimental details}
To train the ESIM model of \citet{Chen2017EnhancedLF}, the decomposable attention model of \citet{Parikh2016ADA}, the BiDAF model of \citet{Seo2016BidirectionalAF}, and the QANet model of \citet{Yu2018QANetCL}, we use the implementations in AllenNLP \citep{allennlp:2018:nlposs}. The models are trained with the same hyperparameters as described in their respective papers.

For each training dataset size, we tune the learning rate on the \textit{original} development set accuracy; the learning rate is halved whenever validation performance ($F_1$ for SQuAD, accuracy for NLI) does not improve, and we employ early stopping with a patience of 5. This ensures that we are not implicitly using additional challenge dataset examples. For each model and amount of challenge dataset examples used for fine-tuning, the reported challenge dataset performance is the performance of the learning rate configuration that yields the best challenge dataset performance. We leave all other hyperparameters (such as the batch size and choice of optimizer) unchanged from the model's original training procedure.

For the Adversarial SQuAD experiments, we experiment with learning rates of 0.00001, 0.0001, 0.001 and 0.01. For the NLI stress test experiments, we experiment with learning rates of 0.000001, 0.00001, 0.0001, 0.0004, 0.001, and 0.01.

We use AllenNLP to run our fine-tuning experiments.

\clearpage
\newpage
\onecolumn
\labeledsection{MultiNLI Mismatched Stress Test Results}{mismatched_results}
\begin{figure*}[!htp]
\begin{tabular}{cc}
\textbf{(a) Word Overlap} & \textbf{(b) Negation} \\
\subfloat[Performance of the ESIM and DA models after fine-tuning on a variable number of word overlap challenge dataset examples generated from the MultiNLI mismatched development set.]{\includegraphics[width=0.45\linewidth]{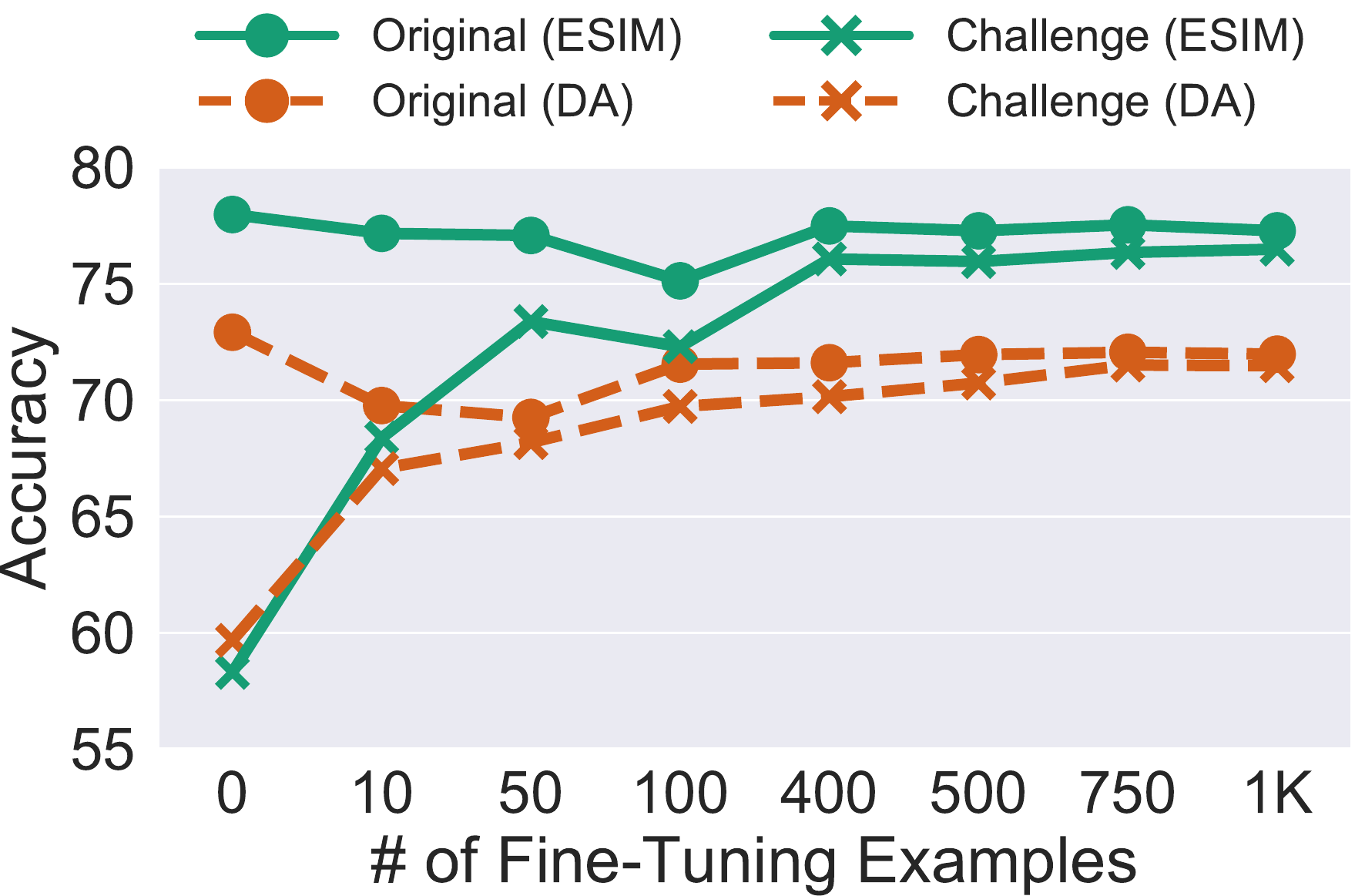}} &
\subfloat[Performance of the ESIM and DA models after fine-tuning on a variable number of negation challenge dataset examples generated from the MultiNLI mismatched development set.]{\includegraphics[width=0.45\linewidth]{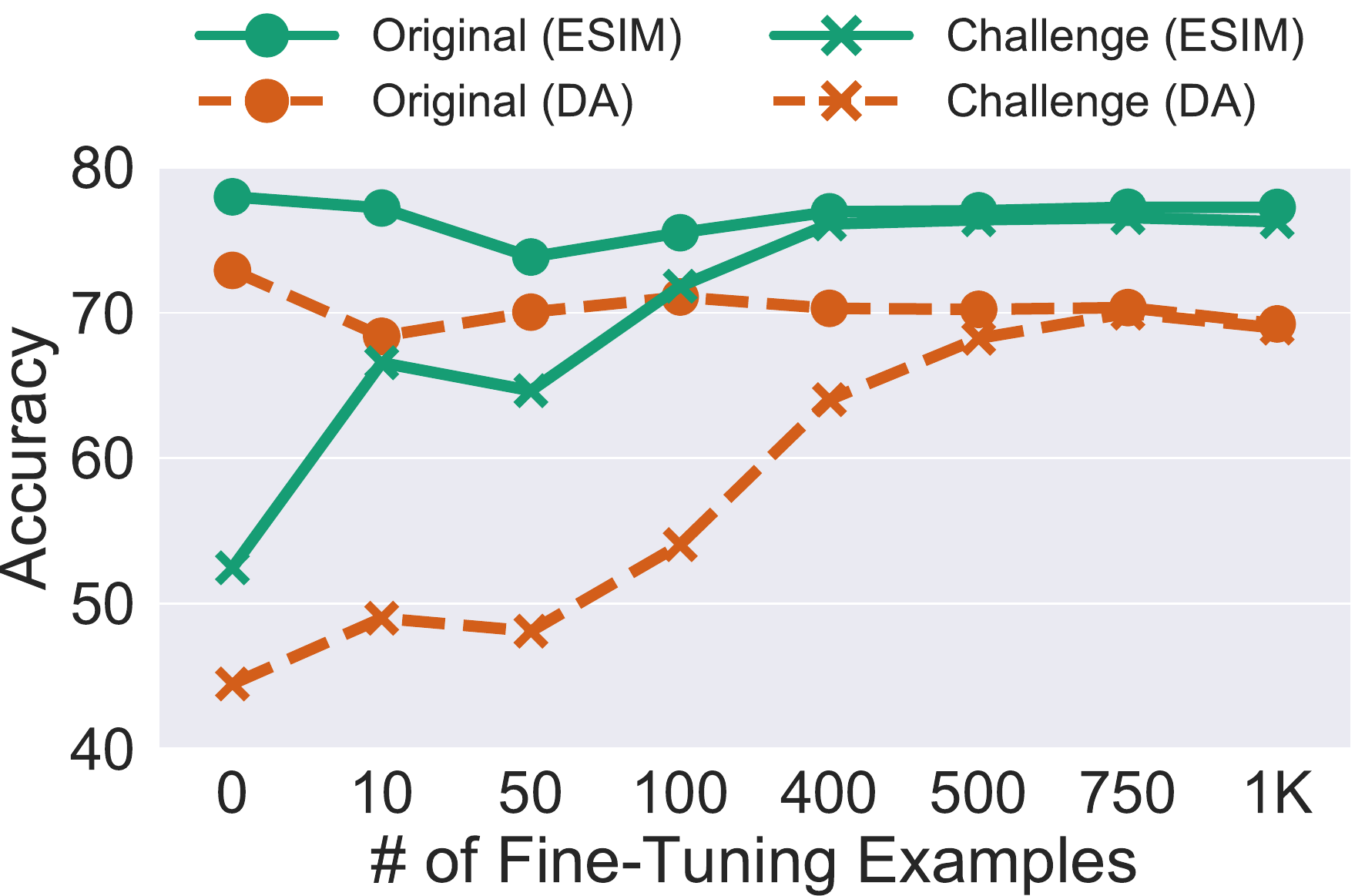}} \\ \\
\textbf{(c) Spelling Errors} & \textbf{(d) Length Mismatch} \\
\subfloat[Performance of the ESIM (with and without an additional character-level component) and DA models after fine-tuning on a variable number of spelling error challenge dataset examples generated from the MultiNLI mismatched development set.]{\includegraphics[width=0.45\linewidth]{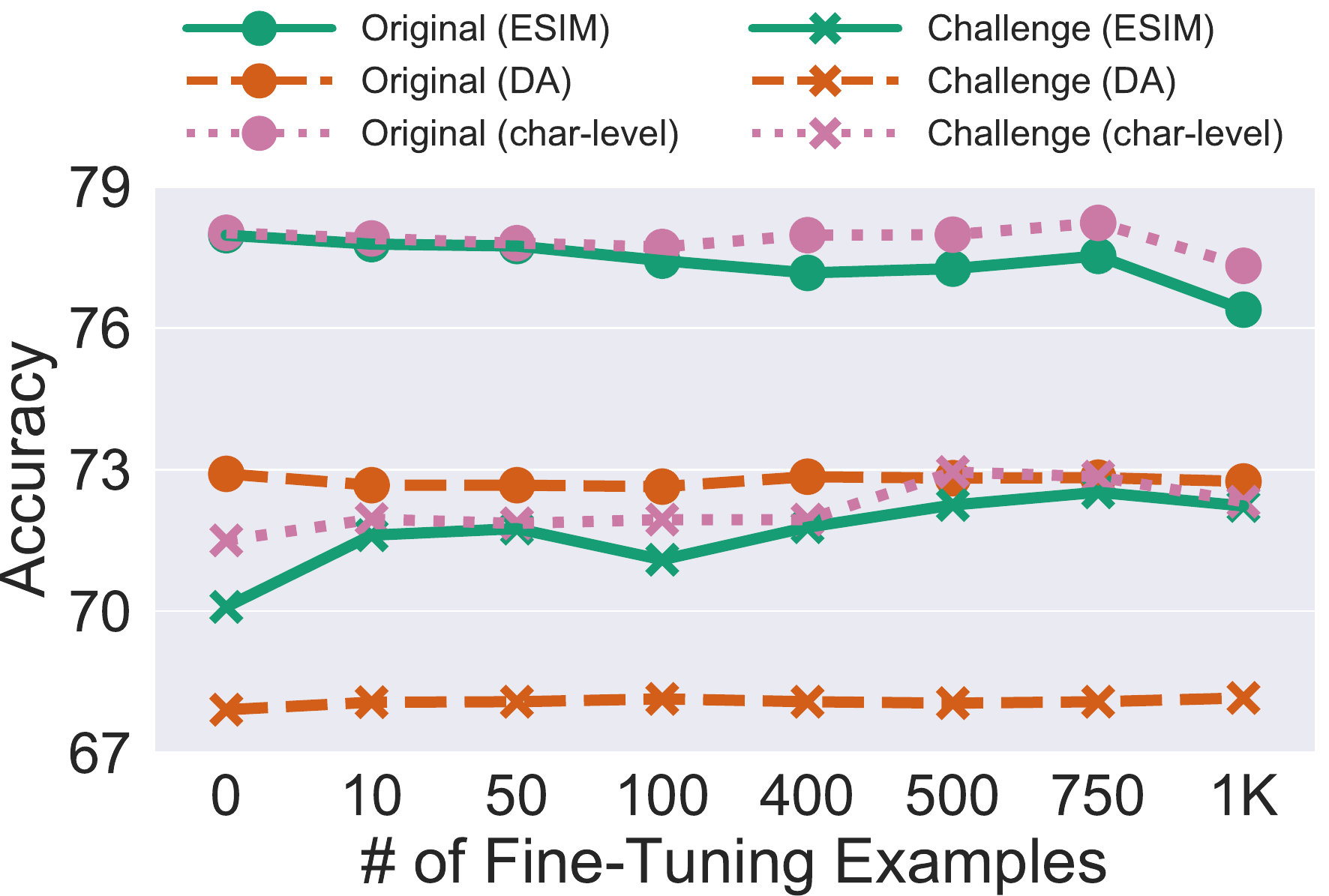}} & \subfloat[Performance of the ESIM and DA models after fine-tuning on a variable number of length mismatch challenge dataset examples generated from the MultiNLI mismatched development set.]{\includegraphics[width=0.45\linewidth]{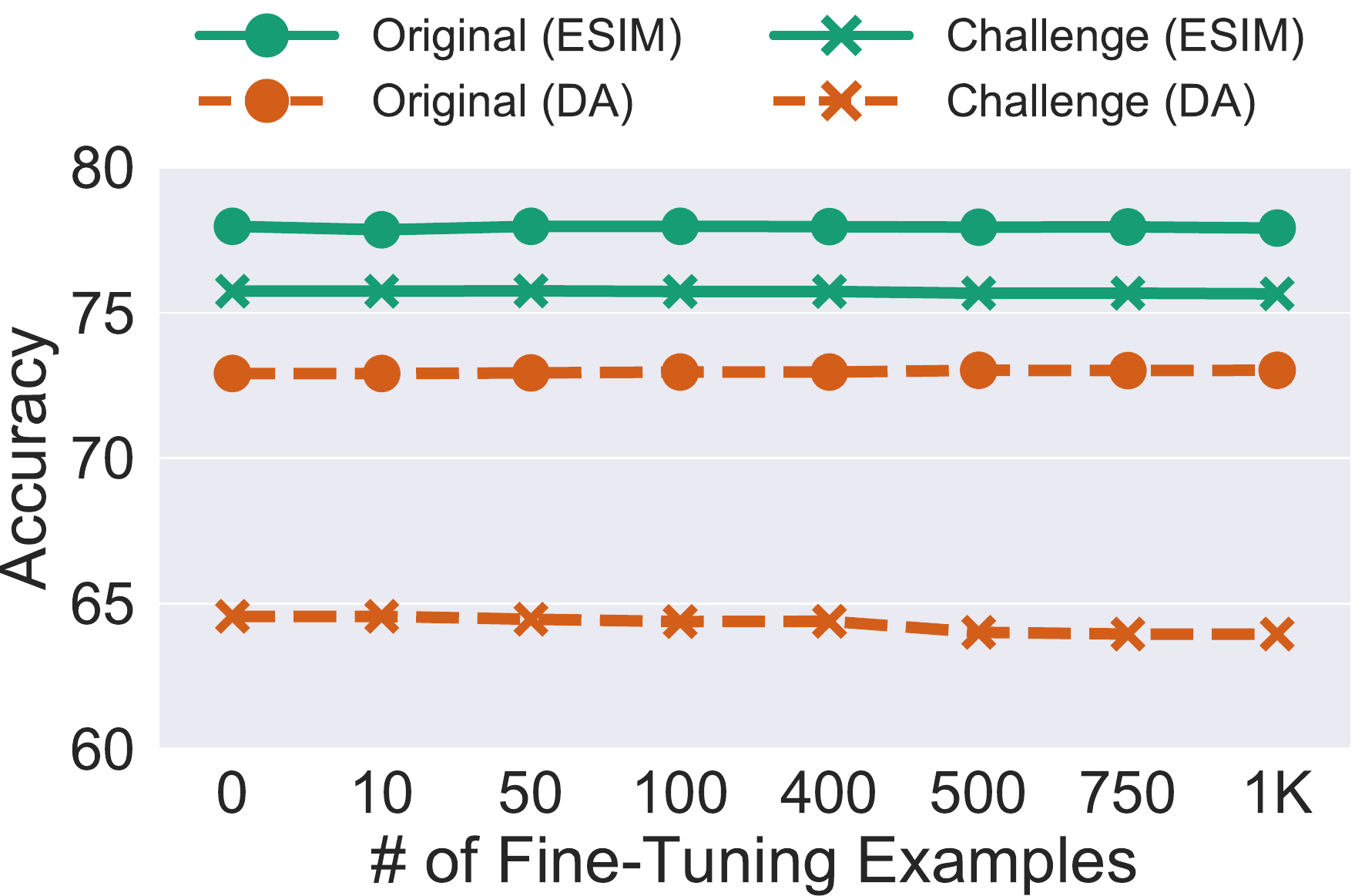}} \\ \\
\textbf{(e) Numerical Reasoning} & \\
\subfloat[Performance of the ESIM and DA models (where the mismatched development set was used during training to control learning rate scheduling and early stopping) after fine-tuning on a variable number of numerical reasoning challenge dataset examples.]{\includegraphics[width=0.45\linewidth]{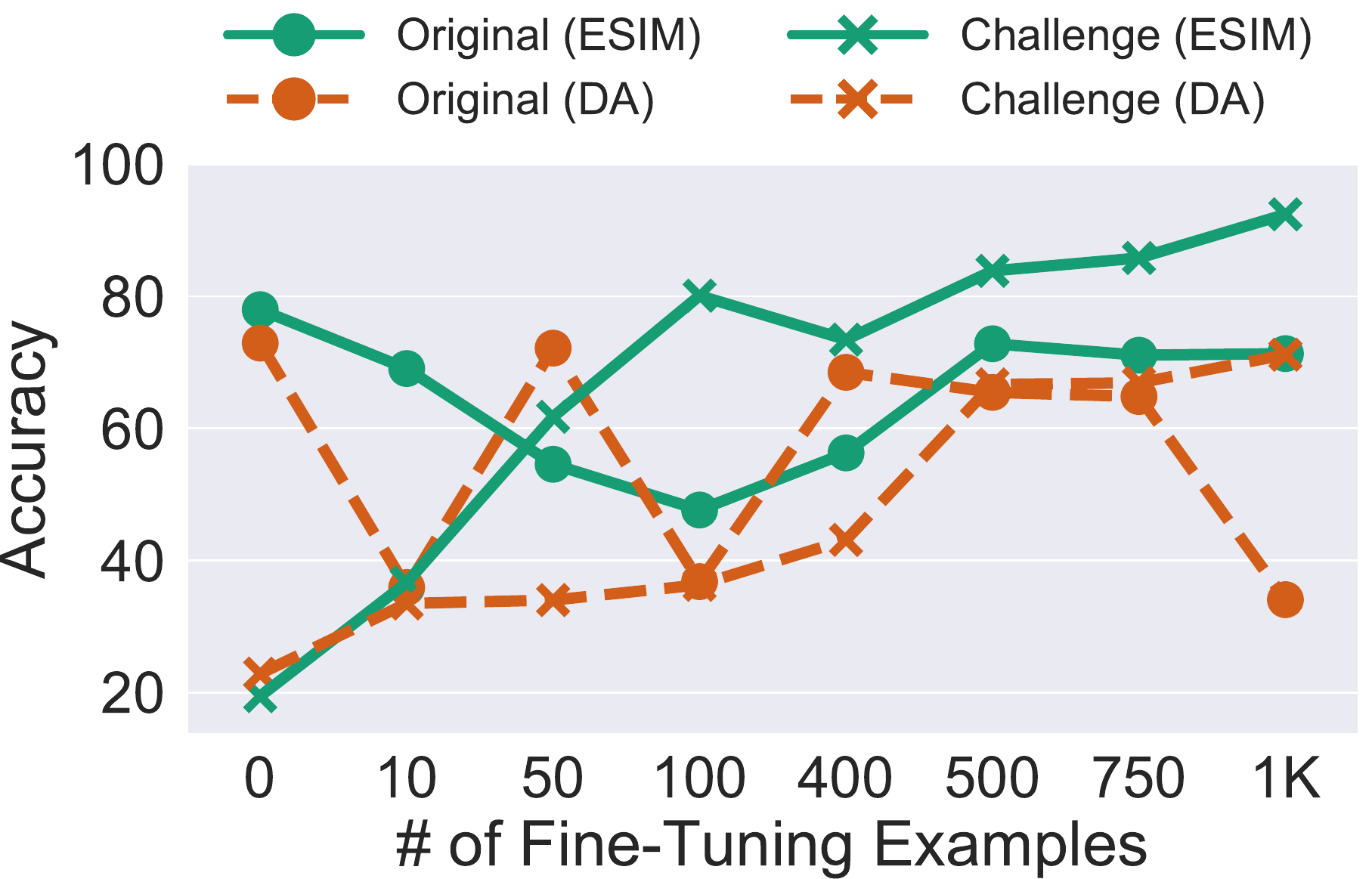}}\\ & \\
\end{tabular}
  \label{fig:all_nli_mismatched_results}
\end{figure*}

\clearpage
\newpage
\twocolumn

\labeledsection{Three Simple Rules for the Numerical Reasoning Dataset}{numeric_explanation}

The numerical reasoning dataset of \citet{Naik2018StressTE} has 7,596 examples in total, with 2,532 in each of the ``entailment'', ``neutral'', and ``contradiction'' categories. With only three rules, we can  correctly classify around 82\% of the examples.

1,235 examples (out of the 7,596 in total) can be correctly labeled as contradiction with the rule: ``more than'' or ``less than'' do not appear in the premise or the hypothesis.

2,664 examples (out of the 6,361 examples left to be considered) contain ``more than'' or ``less than'' in the hypothesis. Of these 2,664 examples, 2,532 have the label ``neutral'', 66 have the label ``entailment'', and 66 have the label ``contradiction''. So, if the hypothesis contains ``more than'' or ``less than'', we predict ``neutral''. This rule leads us to correctly classify 2,532 examples and incorrectly classify 132 examples.

Finally, we have 3,697 examples to be considered. All 3,697 of these examples have ``more than'' or ``less than'' in the premise. 2,466 of these examples are labeled ``entailment'', while 1,231 are labeled ``contradicion''. By assigning 
the label ``entailment'' to examples that contain ``more than'' or ``less than'' in their premise, we correctly classify 2,446 examples and incorrectly classify 1,231 examples.

In total, these three rules result in correct predictions on 6,233 examples out of 7,596 (82.05\%).

\end{appendices}
\end{document}